\documentclass{article}

\PassOptionsToPackage{numbers, compress}{natbib}

\usepackage[final]{neurips_2025}

\usepackage{array}  % 用于扩展表格功能
\newcolumntype{M}[1]{>{\centering\arraybackslash}m{#1}}  % 定义 M{} 列类型
\usepackage{multirow}

\usepackage{graphicx}

\usepackage{float}

\usepackage[utf8]{inputenc} % allow utf-8 input
\usepackage[T1]{fontenc}    % use 8-bit T1 fonts
\usepackage{hyperref}       % hyperlinks
\usepackage{url}            % simple URL typesetting
\usepackage{booktabs}       % professional-quality tables
\usepackage{amsfonts}       % blackboard math symbols
\usepackage{nicefrac}       % compact symbols for 1/2, etc.
\usepackage{microtype}      % microtypography
\usepackage{xcolor}         % colors
\usepackage{tabularx} % 提供X列类型

\title{DeepMath-Creative: A Benchmark for Evaluating Mathematical Creativity of Large Language Models}

\author{%
  An Open-Source Program Initiated by the DeepMath Team\thanks{Founders of DeepMath Team: Xiaoyang Chen, Yuting Gao, Xiang Jiang, Xiangnan Li. Affiliation: School of Mathematical Sciences, Tongji University. GitHub: \url{https://github.com/DeepMathLLM/DeepMath}}  \\
  \textnormal{A Complete List of Contributors is Provided in the Appendix}
}

\begin{document}

\maketitle

\begin{abstract}
To advance the mathematical proficiency of large language models (LLMs), the DeepMath team has launched an open-source initiative aimed at developing an open mathematical LLM and systematically evaluating its mathematical creativity. This paper represents the initial contribution of this initiative. While recent developments in mathematical LLMs have predominantly emphasized reasoning skills, as evidenced by benchmarks on elementary to undergraduate-level mathematical tasks, the creative capabilities of these models have received comparatively little attention, and evaluation datasets remain scarce. To address this gap, we propose an evaluation criteria for mathematical creativity and introduce DeepMath-Creative, a novel, high-quality benchmark comprising constructive problems across algebra, geometry, analysis, and other domains. We conduct a systematic evaluation of mainstream LLMs' creative problem-solving abilities using this dataset. Experimental results show that even under lenient scoring criteria---emphasizing core solution components and disregarding minor inaccuracies, such as small logical gaps, incomplete justifications, or redundant explanations---the best-performing model, O3 Mini, achieves merely 70\% accuracy, primarily on basic undergraduate-level constructive tasks. Performance declines sharply on more complex problems, with models failing to provide substantive strategies for open problems. These findings suggest that, although current LLMs display a degree of constructive proficiency on familiar and lower-difficulty problems, such performance is likely attributable to the recombination of memorized patterns rather than authentic creative insight or novel synthesis.

\end{abstract}

\section{Introduction}

In recent years, large language models (LLMs), represented by GPT~\cite{openai_o3_o4} and DeepSeek~\cite{guo2025deepseek}, have demonstrated impressive reasoning capabilities in mathematical problem-solving. Empirical studies have shown that LLMs exhibit proficiency in fundamental arithmetic and multi-step logical reasoning. For instance, the Multi-LogiEval benchmark reports an average accuracy of 68\% for LLMs on multi-step logical reasoning tasks, sustaining a performance level of approximately 43\% on reasoning tasks of depth-5~\cite{patel2024multi}. Furthermore, LLMs have shown considerable promise as middle school mathematics tutoring assistants across three key tasks: hint generation, comprehensive solution provision, and exercise creation~\cite{ramanathan2024comparison}. Notably, the AlphaGeometry system successfully solved 25 out of 30 International Mathematical Olympiad (IMO) geometry problems, approaching the level of an average IMO gold medallist~\cite{trinh2024solving}.
However, current datasets predominantly emphasize basic or competition-style problems, with limited attention directed towards cutting-edge, creative, and exploratory mathematical challenges. Consequently, these datasets do not adequately reflect the true extent of mathematical creativity exhibited by large language models.

A fundamental inquiry arises: how can mathematical creativity be effectively evaluated? To address this, we propose a framework based on three key dimensions. The first and arguably most profound dimension is the generation of novel concepts. The introduction of new concepts and ideas signifies a qualitative leap in mathematical understanding. For instance, the advent of Riemannian metrics laid the groundwork for modern differential geometry and general relativity; the concept of differentiable manifolds enabled the widespread application of local linearization techniques in mathematical and physical models of spacetime; the genesis of group theory revolutionized structural thinking in algebra; and the notion of topological spaces transcended the limitations imposed by traditional distance-based frameworks. These foundational concepts have served as enduring catalysts for the advancement of modern mathematics. The second dimension lies in the invention of novel methods. New methodologies furnish potent tools and pathways for addressing intricate problems. For example, the theory of generalized functions enabled a departure from classical function frameworks, leading to breakthroughs in fields such as quantum mechanics. The calculus of variations emerged as a pivotal instrument in the study of geometric optimization, while Bochner's technique marked a significant milestone in geometric analysis. The third dimension involves the creation of novel examples. The construction of counterexamples plays a critical role in mathematical inquiry, serving to test the universality of propositions and delineate the boundaries of theoretical constructs, thereby propelling the development of new theories and the refinement of existing ones. For example, Milnor's discovery of exotic spheres~\cite{milnor1956exotic} overturned the established understanding of high-dimensional sphere structures and inaugurated a new era in exotic manifold theory and differential topology. Weierstrass's construction~\cite{weierstrass1872} of a function continuous everywhere but differentiable nowhere directly spurred the development of real analysis. The existence of such examples not only continually fortifies the rigor of theoretical frameworks but also stands as a crucial manifestation of mathematical creativity. Consequently, the creation of novel counterexamples holds substantial theoretical significance and serves as an indispensable driving force for mathematical progress.

The historical processes underlying the creation of novel mathematical concepts and methods by humans are frequently unstructured and contingent. Consequently, reconstructing the cognitive trajectories and exploratory processes of mathematicians through large-scale, standardized datasets proves exceedingly difficult, and relevant systematic process data remain exceptionally scarce. Moreover, while large language models can generate outputs that appear to be novel mathematical concepts and methods, discerning whether this "novelty" represents genuine originality or merely a recombination or variation of existing knowledge presents a significant challenge due to the current limitations of evaluation mechanisms. Notably, the process of proving or refuting mathematical propositions often necessitates the construction of novel mathematical examples. In contrast to abstract concepts and theories, these concrete, constructible examples are more amenable to verification and thus offer a valuable entry point for evaluating the creative capabilities of large language models.

To address the current absence of a dedicated dataset for evaluating the mathematical creativity of large language models, we have constructed DeepMath-Creative, a high-quality benchmark encompassing several important branches of mathematics, including algebra, topology, geometry, and analysis. This benchmark features a challenging set of innovative problems, comprising two types of inquiry-based questions: (1) problem requiring a formal proof, which necessitate the model to construct a mathematical object to validate a given proposition; and (2) problem requiring a counterexample, which require the model to construct a counterexample to invalidate a given proposition. Collectively, these two problem types provide a comprehensive framework for the assessment of the model’s mathematical creativity. Subsequently, we employ a rigorous experimental procedure and well-defined evaluation criteria to systematically assess several mainstream large language models, including the GPT and DeepSeek series. Our experimental findings reveal that when confronted with creative mathematical problems, existing models still exhibit limitations such as misdirection in construction, flawed reasoning, unnecessarily verbose solutions, and a lack of convergence towards correct solutions. In future work, we intend to explore reinforcement learning for training and release the DeepMath-Creative Model to continuously enhance the creative mathematical abilities of large language models.

In parallel, we have also developed an evaluation set comprising 170 foundational problems, spanning key branches such as mathematical analysis, advanced algebra, probability theory, mathematical statistics, combinatorics, complex analysis, number theory, partial differential equations, and operations research. Evaluation results obtained from this set indicate that large language models demonstrate robust reasoning abilities in solving undergraduate-level mathematical problems, achieving an overall accuracy of approximately 85\%. This level of performance suggests that current models have attained a significant proficiency in fundamental reasoning tasks. However, existing foundational problem sets prove inadequate for evaluating creative capabilities, underscoring the pressing need to design more innovative and open problem sets to comprehensively assess models' creative constructive abilities.

Furthermore, we have compiled a set of open problems. Initial evaluation results, however, reveal that current models have yet to provide information of substantial referential value for advanced mathematical research. The application of large language models to cutting-edge mathematical research necessitates further exploration of strategies to significantly enhance their mathematical creativity.

Experimental results indicate that current large language models still face considerable challenges in mathematical creativity tasks, including misjudging solution strategies, exhibiting weak logical reasoning, and producing overly lengthy and unfocused derivations. Despite the use of highly lenient evaluation criteria---where only key solution components were assessed and minor errors were disregarded---the best-performing model, O3 Mini, achieved an accuracy of only around 70\%. Notably, the majority of the benchmark problems are at the undergraduate level and primarily test basic constructive thinking. However, the model's performance deteriorates significantly on more difficult problems, and it fails to provide effective strategies when faced with open problems. These observations suggest that while current models demonstrate some level of constructive capability on known, lower-difficulty problems, their performance is likely driven by recombination of memorized knowledge rather than genuine creativity. This underscores the persistent limitations of existing models in achieving true mathematical creativity.

\section{Related Work}

Numerous studies in the field of mathematical reasoning have focused on constructing high-quality datasets to facilitate algorithmic optimization and enhance the capabilities of large language models. Representative works in this area include:

\textbf{GSM8K}~\cite{cobbe2021gsm8k} is a dataset consisting primarily of basic arithmetic and word problems designed for elementary and secondary school levels. It is intended to evaluate models' multi-step reasoning and language understanding abilities.

\textbf{MATH}~\cite{hendrycks2021math} encompasses problems ranging from middle school to the International Mathematical Olympiad (IMO) level. It assesses models' performance on complex mathematical reasoning and comprehensive problem-solving tasks.

\textbf{AIME2024}~\cite{maxwell2024aime} is derived from actual questions from the American Invitational Mathematics Examination (AIME). It emphasizes rigorous reasoning and deductive skills in algebra, geometry, and number theory, focusing on integer-based solutions.

\textbf{MMLU (Math)}~\cite{hendrycks2021mmlu} represents the mathematics subset of the Massive Multitask Language Understanding (MMLU) benchmark. It includes four levels: abstract algebra, college mathematics, high school mathematics, and elementary mathematics, with all problems formatted as multiple-choice questions.

\textbf{FrontierMath}~\cite{glazer2024frontiermath} targets advanced mathematical problems at the graduate level and beyond, covering modern mathematical fields such as number theory, algebraic geometry, and category theory. It evaluates models' abilities in abstract concept comprehension, structured reasoning, and original problem-solving.

While these datasets have contributed significantly to advancing mathematical reasoning in language models, they share several limitations. First, their problem formats are relatively traditional, emphasizing computational accuracy and logical completeness, while offering limited evaluation of models' creativity, counterexample construction, and original reasoning. Second, the mathematical domains are predominantly rooted in basic education or competition-style content, lacking in highly abstract and innovation-driven professional topics. Moreover, many state-of-the-art models have reached near-saturation performance on these benchmarks, diminishing their effectiveness in differentiating model capabilities. To address these issues, we introduce a high-quality benchmark that emphasizes creativity in professional mathematical domains. Using this benchmark, we systematically evaluate and compare the creative problem-solving performance of contemporary large language models.

\section{Construction of the Benchmark}

\subsection{Design Principles}
To accurately evaluate the mathematical creativity of large language models, the benchmark design adheres to the fundamental principle of innovative constructiveness. The developed dataset focuses on core branches of mathematics, including algebra, topology, real analysis, and geometry, representing significant directions and inherent challenges in mathematical research. The selected problems emphasize creative construction, aiming to assess whether models can transcend mere memorization, engage in independent problem exploration, and generate innovative solutions. Through this design, we seek to evaluate the capacity of models to exhibit independent thinking and creativity when confronted with previously unseen problems.

As famously articulated by Poincaré~\cite{poincare1905value}: ``What is mathematical creation? It does not consist in making new combinations with mathematical entities already known. Anyone could do that, but the combinations so made would be infinite in number and most of them absolutely without interest. To create consists precisely in not making useless combinations and in making those which are useful and which are only a small minority. Invention is discernment, choice.'' Within the context of undergraduate and master's-level mathematics, we posit that creativity frequently manifests as constructiveness---specifically, through the construction of mathematical objects, functions, or structures that satisfy specific properties to prove or disprove a given proposition. Consider, for instance, a problem from real analysis~\cite{cheng2004counterexamples}: Let $\{g_n\}$ be a uniformly bounded sequence of measurable functions, and let $p > 1$. Suppose that for any measurable function $f$ with integrable $|f|^p$, the following holds:
\[
\lim_{n \to \infty} \int_a^b f g_n \, dx = \int_a^b f g \, dx,
\]
does it necessarily follow that $\{g_n\}$ converges to $g$ in measure? If the proposition holds, provide a proof; if not, construct a counterexample. The conclusion of this problem is negative, and its solution necessitates the construction of a specific sequence $\{g_n\}$ to demonstrate the lack of convergence to $g$ in measure. Mathematical results derived solely through the combination and deduction of known conclusions do not inherently reflect mathematical creativity. Mathematical work transcends a mechanical and redundant exercise in arbitrary combinations; rather, the creator must judiciously select and connect the limited number of useful combinations, engaging in deliberate effort that enables subconscious creativity to flourish. What, then, constitutes a creative mathematical problem? We argue that a problem exhibits significant creative characteristics if its solution hinges upon the construction of an example to prove or refute a proposition and if it demands the model to comprehend the problem and explore potential solutions independently, without reliance on a predetermined solution path. An in-depth assessment of the model’s ability in such constructive tasks can thus be achieved, aligning with the third dimension of mathematical creativity outlined earlier.

\begin{figure}[H]
  \centering
  \includegraphics[width=0.8\textwidth]{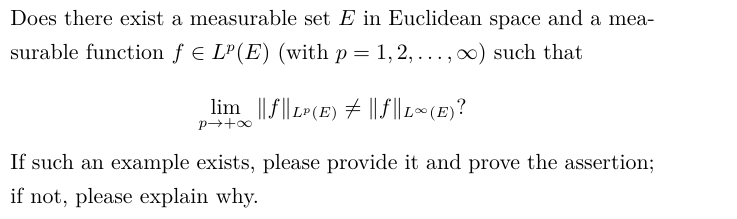}
  \caption{Example of an innovative mathematical problem structure designed for the model evaluation process.}
  \label{fig:problem}
\end{figure}

For the problems within the innovative dataset, we introduce a novel and challenging problem formulation. The design employs a unified structure for all problems, presented in the format: ``Problem description + If the proposition holds, please prove it; if the proposition does not hold, please provide a counterexample.'' This format not only exhibits clarity and explicitness but also departs from the traditional unidirectional question format that elicits only a definitive answer. It encourages the model to actively engage in comprehensive logical analysis and multidimensional thinking prior to formulating a response, thereby facilitating a more authentic assessment of the model’s creativity. Overall, this bidirectional inquiry-based format substantially broadens the model’s cognitive boundaries, permitting a more thorough reflection of its performance on higher-order tasks and providing robust support for the quantitative evaluation of mathematical creativity. Figure~\ref{fig:problem} presents a mathematical problem we designed with an innovative and open structure.

\subsection{Data Collection}
To ensure the high quality and originality of the benchmark, all mathematical problems utilized in this study were meticulously designed and annotated by experts in the field of mathematics. The data collection and design team comprised professors and graduate students from the Department of Mathematics. Drawing upon their research expertise and professional knowledge, as well as referencing established professional mathematics textbooks, they creatively formulated problems and devised unique problem formats. The design and collection process adhered to a multi-round expert discussion and review methodology, rigorously ensuring a high standard of problem provenance. This process guaranteed that the problems achieved exceptionally high standards in logical rigor, mathematical correctness, and accuracy of problem descriptions, thereby ensuring the overall quality and credibility of the dataset.

Through this rigorous process, we constructed an evaluation set containing 179 problems. The innovative dataset is distributed across the following mathematical domains: approximately 50\% in algebra, 15\% in topology, and 35\% in analysis. Regarding difficulty levels, around 60\% of the problems correspond to undergraduate-level mathematics, while approximately 40\% correspond to master's-level mathematics. Furthermore, the dataset comprises two categories of problems: those requiring formal proofs (approximately 40\% of the dataset) and those demanding counterexamples (about 60\%).  These are designed to respectively evaluate the language models' abilities in original reasoning and critical thinking.

\subsection{Evaluation Metrics and Procedure}
To comprehensively evaluate the performance of language models on mathematical creativity tasks, we devised a systematic evaluation framework integrating both quantitative and qualitative metrics. For the quantitative assessment, the innovative dataset employs two indicators: ``direction accuracy'' and ``process accuracy,'' which respectively assess the correctness of the answer's overall direction and the accuracy of the solution process, providing an objective quantification of model performance. For the qualitative assessment, all outputs are evaluated by a team of mathematical experts through manual grading, focusing on the logical rigor, clarity of mathematical expression, and originality of the model's solutions, thereby capturing subtle nuances in model performance on creative problems.

This innovative dataset comprises problems requiring a formal proof and problems requiring a counterexample. The detailed scoring criteria for these two types of problems are presented in Table~\ref{tab:results}. For problems requiring a formal proof: if the model mistakenly treats the proposition as false and attempts to refute it by constructing a counterexample---thereby selecting an incorrect problem-solving direction---it receives a score of 0. If the model chooses to prove the proposition but the reasoning process contains significant flaws or omits key steps, resulting in an incomplete proof, it receives a score of 0.5. If the model provides a complete and rigorous proof, with clear logical reasoning and no major omissions, it receives a score of 1. For problems requiring a counterexample: if the model erroneously attempts to prove the proposition true---thereby selecting an incorrect problem-solving direction---it receives a score of 0. If the model recognizes the proposition as false and attempts to construct a counterexample, but the counterexample contains logical flaws or fails to satisfy the problem's conditions, it receives a score of 0.5. If the model successfully identifies the proposition as false and constructs a logically sound counterexample that satisfies the conditions, it receives a score of 1 (if the model provides multiple examples, a score of 1 is awarded as long as at least one correct counterexample is given). This scoring system focuses on evaluating the model’s abilities in selecting the correct problem-solving direction and constructing valid examples. It assesses both the model’s judgment of the proposition and its mathematical creativity. This approach not only facilitates the differentiation of model performance in task understanding and creative construction but also provides a quantitative basis for subsequent result analysis.

\begin{table}
  \caption{Scoring criteria for mathematical problems}
  \label{tab:results}
  \centering
  \begin{tabular}{>{\centering\arraybackslash}m{3.2cm} 
                  >{\centering\arraybackslash}m{4.2cm} 
                  >{\centering\arraybackslash}m{4.2cm} 
                  >{\centering\arraybackslash}m{1.2cm}}
    \toprule
    Problem type & Model response & Evaluation criteria & Score \\
    \midrule
    \multirow{3}{3.2cm}{Problem requiring a formal proof} 
      & Constructs a counterexample & Incorrect direction of construction & 0 \\
      & Attempts a proof with flaws & Partial logic, missing or incorrect steps & 0.5 \\
      & Provides a complete correct proof & Logical, rigorous, and complete proof & 1 \\
    \midrule
    \multirow{3}{3.2cm}{Problem requiring a counterexample} 
      & Attempts to prove the statement & Incorrect direction of construction & 0 \\
      & Constructs flawed counterexample & Reasonable attempt, but violates some condition & 0.5 \\
      & Constructs a valid counterexample & Fully correct, satisfies all conditions & 1 \\
    \bottomrule
  \end{tabular}
\end{table}

The evaluation experiments were conducted in strict accordance with a standardized procedure. All participating language models were uniformly integrated into the evaluation system via API interfaces. Each model autonomously generated responses based on a standardized prompt format. The generated responses were subsequently subjected to manual review and scoring according to the established scoring guidelines, ensuring objectivity, fairness, and accuracy in the evaluation process. Ultimately, the evaluation results clearly delineate the performance differences among the models on creative mathematical tasks and provide a reliable basis for further model optimization and development.

\section{Experimental Results and Analysis}
This chapter presents a systematic analysis of mainstream models based on the DeepMath-Creative benchmark constructed in the preceding sections. The experiments focus on evaluating model performance along two key dimensions: problem-solving direction and creative construction ability. In total, five current mainstream models were selected for comparative evaluation, specifically: GPT o3-mini (2025-01-31 version), Claude-3-7-Sonnet (2025-02-19 version), Gemini-2.0-Flash, DeepSeek R1, and Qwen QwQ-32B. The selection of these models reflects a balance of varying model scales, pretraining strategies, and architectural designs, capturing the technological diversity of the current language modeling landscape. This diversity ensures that the experimental conclusions possess broad applicability and significant academic reference value. All models were evaluated under a unified testing environment with consistent input formats, ensuring fairness and reproducibility of the experiments.

\subsection{Overview of Experimental Results}
This section provides an overall analysis of the models' outputs in the experiments. Based on the evaluation criteria, the problems are categorized according to the models' scores, and the observed patterns are summarized and analyzed.

\subsubsection{Model Output Analysis}
For the outputs generated by different models, we obtained the following specific observations and conclusions:

\begin{enumerate}
    \item \textbf{GPT O3-mini}: The outputs of this model often exhibit vague structure and expression, particularly on complex or highly open problems, where the clarity of logic and definiteness of conclusions are insufficient.

    \item \textbf{Claude-3-7-Sonnet}: This model displays a pronounced bipolar characteristic. For problems it handles well, its answers are typically concise, accurate, and logically rigorous; however, for problems it has not sufficiently mastered, its answers tend to contain significant errors or logical flaws.

    \item \textbf{Gemini 2.0 Flash}: The overall performance is moderate, with some details showing reasonable rigor and clear output formatting. However, notable omissions frequently occur in the final answer section of many problems.

    \item \textbf{DeepSeek R1}: Although the overall performance is stable, some responses demonstrate spurious proof behavior, in which the model presents an incorrect construction and arrives at the correct conclusion via erroneous intermediate steps. Another clear tendency of this model is that for problems it has mastered, the answers are often concise and straightforward; conversely, for problems where the output is lengthy and involves repeated reasoning, the final answers are typically incorrect.

    \item \textbf{Qwen QwQ-32B}: This model tends to provide excessively detailed reasoning and construction processes, but the overly lengthy outputs frequently exceed the text box limits. In many cases, the model failed to provide a clear final conclusion, instead becoming trapped in continuous and repetitive reasoning loops, thereby undermining the validity of the evaluation.
\end{enumerate}

\subsubsection{Feature Analysis by Problem Score}
To conduct a detailed analysis of the models' creative constructive abilities in mathematics, we analyzed the main features and shortcomings of the model outputs based on the evaluation criteria. As shown in Table~\ref{tab:model-results}, for problems scored between 0 and 0.5, the models' outputs generally exhibited the following typical deficiencies: incorrect problem-solving direction, construction processes with obvious flaws and spurious proof phenomena, and overly verbose outputs lacking a clear final conclusion. In contrast, for problems scored 1 point, the models' outputs typically displayed the following characteristics: a concise and rigorous construction process, accurate use of key terminology and methods, and a clear stopping point in the reasoning process.

\begin{table}[ht]
  \caption{Performance and common characteristics of different models}
  \label{tab:model-results}
  \centering
  \begin{tabular}{M{2.5cm} M{1.2cm} M{1.2cm} M{1.2cm} M{1.2cm} M{1.2cm} M{5cm}}
    \toprule
    \multirow{2}{*}{Score category} 
      & \multicolumn{5}{c}{Number of problems for different models} 
      & \multirow{2}{*}{\centering Common characteristics} \\
    \cmidrule(lr){2-6}
      & GPT & Claude & Gemini & DeepSeek & Qwen & \\
    \midrule
    0--0.5 score 
      & 46 & 80 & 65 & 51 & 86 
      & Incorrect direction, spurious reasoning process, no clear conclusion \\
    1 score 
      & 133 & 99 & 114 & 128 & 93 
      & Concise and rigorous process, accurate method application, clear stopping point in reasoning \\
    \bottomrule
  \end{tabular}
\end{table}

Our experimental results further show that, despite the adoption of lenient scoring criteria---focusing solely on key solution elements while overlooking minor errors---the best-performing model, O3 Mini, achieved only 70\% accuracy. Notably, this performance was primarily attained on undergraduate-level problems that required basic constructive reasoning. As task difficulty increased, the model's accuracy declined significantly, and it consistently failed to propose meaningful strategies for open problems. These findings indicate that while current models demonstrate some degree of constructive ability on familiar and relatively simple tasks, such ability is likely rooted in the recombination of memorized patterns rather than in genuine creativity.

\subsection{Analysis of representative cases}

Building on the overall analysis presented in the preceding section, this subsection conducts a more in-depth examination by incorporating representative cases, revealing several common patterns and issues across the models.

\begin{enumerate}

\item \textbf{Wrong direction.} In problems explicitly requiring a direct proof, some models mistakenly interpreted the proposition as false and instead attempted to construct a counterexample. Conversely, in tasks that demand providing a counterexample or refuting a given statement, certain models incorrectly proceeded with a direct proof. This directional bias reflects a significant weakness in the models' ability to accurately interpret the intended problem requirements. For example, Figure~\ref{fig:claude-response} illustrates a case where the correct solution involves constructing a counterexample, but Claude mistakenly attempted to prove the statement instead.

\begin{figure}[H]
\centering
\includegraphics[width=0.8\textwidth]{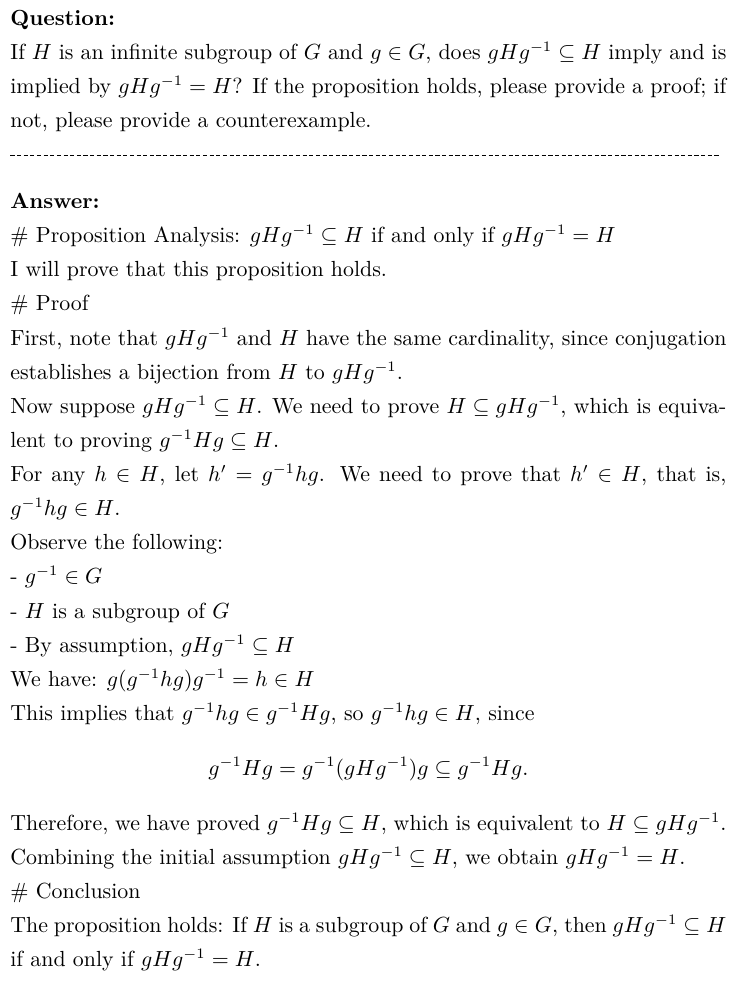}
\caption{Claude's Response: Attempting a Proof in a Counterexample Task}
\label{fig:claude-response}
\end{figure}

\item\textbf{Defective Problem-Solving Process.} Although the model generally follows the correct direction, its problem-solving process still exhibits notable shortcomings. For instance, in tasks requiring the construction of counterexamples, the model is able to approach the problem from the correct perspective but often produces counterexamples that contain factual errors. As illustrated in Figure~\ref{fig:DeepSeek-example}, the counterexample generated by DeepSeek incorrectly asserts that ``\( X \times X \) and \( Y \times Y \) are isometric.'' Furthermore, when the model generates excessively lengthy and repetitive outputs with circular and unfocused content, it typically indicates that the problem exceeds the model's knowledge base or cognitive capacity. Consequently, such responses often fail to yield a complete and correct solution. Figure~\ref{fig:Qwen-response} presents an example from Qwen, where the response is verbose and redundant, ultimately failing to provide a definitive answer, with only partial information shown in the figure.

\begin{figure}[H]
\centering
\includegraphics[width=0.75\textwidth]{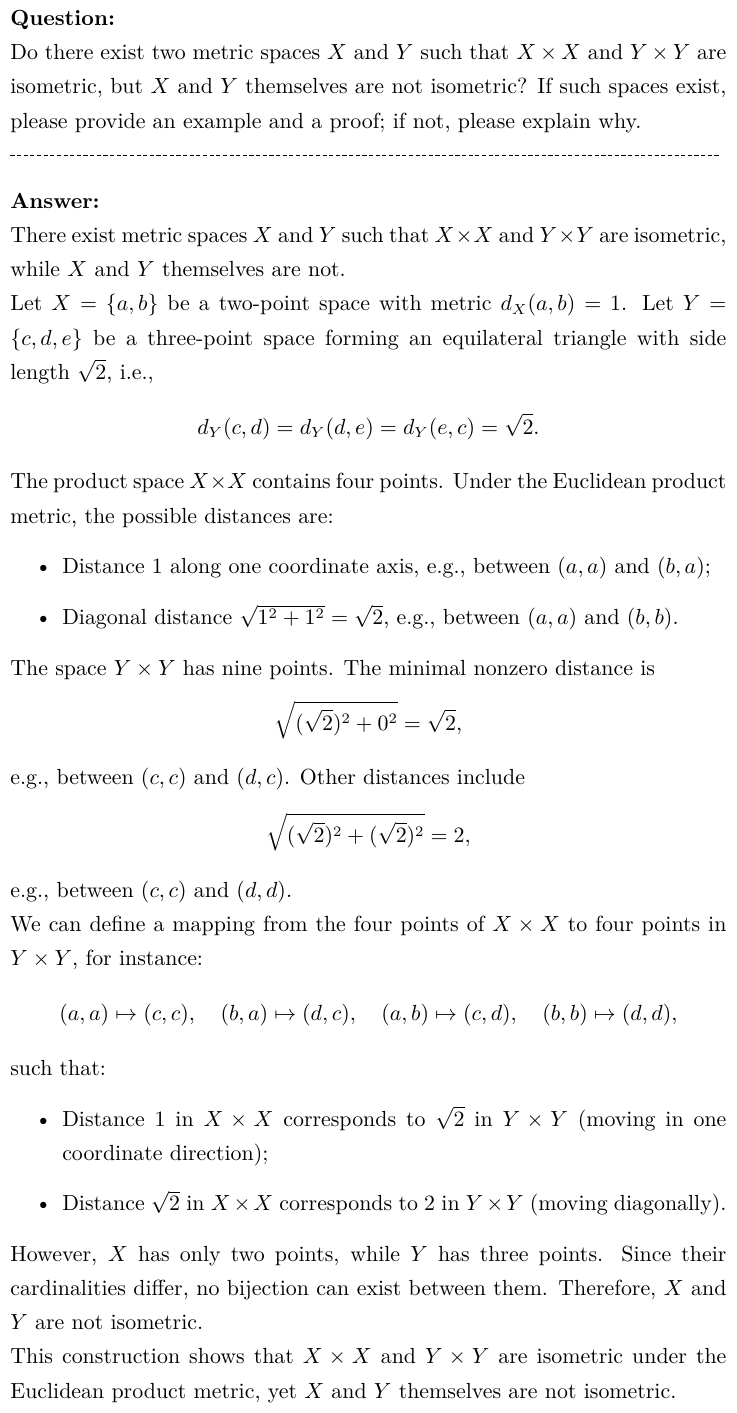}
\caption{DeepSeek's Response: Incorrect Counterexample}
\label{fig:DeepSeek-example}
\end{figure}

\begin{figure}[H]
\centering
\includegraphics[width=0.8\textwidth]{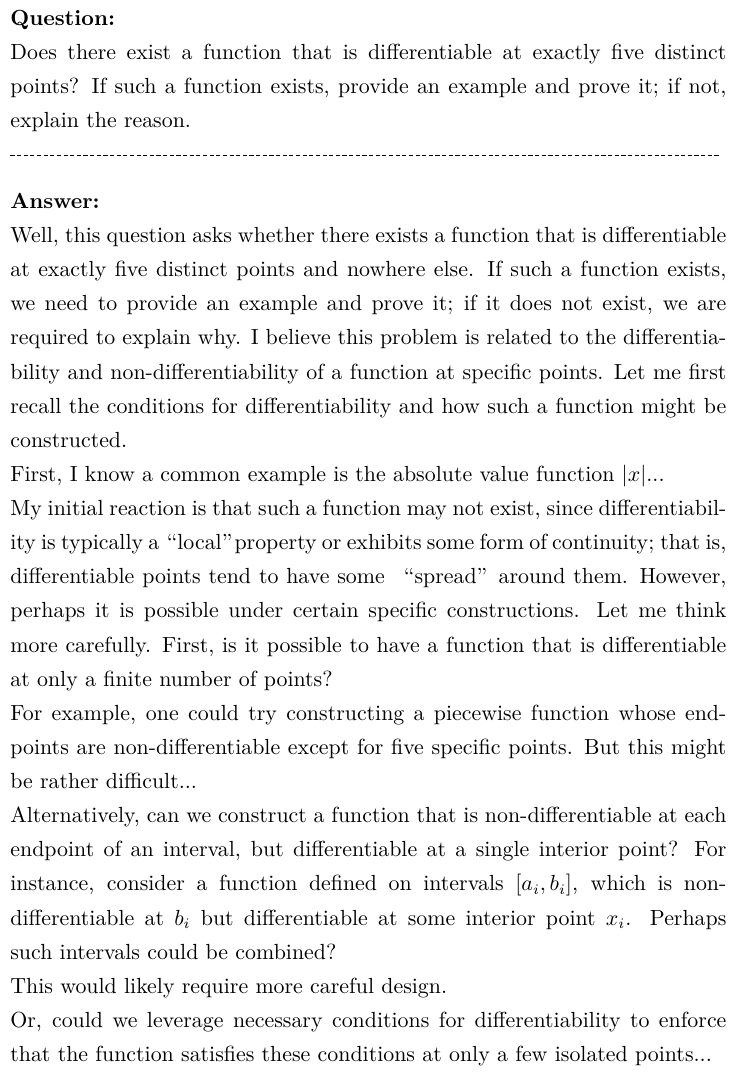}
\caption{Qwen's Response: Lengthy but Incomplete Solution}
\label{fig:Qwen-response}
\end{figure}

\end{enumerate}

In summary, the analysis presented in this section provides a detailed account of the characteristics and shortcomings exhibited by the models during testing, which can inform future improvements and optimizations of such models.

\section{Discussion}
Currently, large language models have achieved a relatively high level of proficiency in solving elementary mathematical problems, demonstrating the ability to provide accurate and well-structured reasoning. However, systematic evaluation based on the DeepMath-Creative benchmark indicates that when confronted with creative mathematical problems, existing models still exhibit issues such as incorrect construction directions, fallacious justifications, and overly lengthy, non-convergent responses. It is worth noting that all mathematical problems within the DeepMath-Creative benchmark possess standard answers. Nevertheless, the creative and constructive abilities of large language models have not yet reached an ideal level, revealing their limitations when tackling genuinely open mathematical problems. Experimental results demonstrate that current large models remain significantly immature in mathematical creativity. To address this persistent challenge, we plan to further employ reinforcement learning for training and release a dedicated model, the DeepMath-Creative Model, with the aim of continuously enhancing the mathematical creativity of large language models.

\newpage

\bibliographystyle{plainnat}  % NeurIPS默认的 natbib 样式
\bibliography{reference}     % 指向你的 .bib 文件名（无 .bib 后缀）

%%%%%%%%%%%%%%%%%%%%%%%%%%%%%%%%%%%%%%%%%%%%%%%%%%%%%%%%%%%%

\appendix

\section{Appendix. List of Contributors and Affiliations}
The complete list of contributors and their affiliations is presented below (in alphabetical order by surname):

\begin{itemize}
    
    \item Xiaoyang Chen, \textit{Tongji University}

    \item Xinan Dai, \textit{Fudan University}
    
    \item Yu Du, 
    \textit{Tianjin University}
    
    \item Qian Feng%, 
    %\textit{ }
    
    \item Naixu Guo%, 
    %\textit{ }
    
    \item Tingshuo Gu, \textit{University of Hong Kong}
    
    \item Yuting Gao, \textit{Tongji University}

    \item Yingyi Gao, \textit{University of Warwick}

    \item Xudong Han, \textit{MBZUAI}, \textit{LibrAI}

    \item Xiang Jiang, \textit{Tongji University}
    
    \item Yilin Jin, \textit{Tongji University}
    
    \item Hongyi Lin, \textit{Tongji University}
    
    \item Shisheng Lin, \textit{Tongji University}
    
    \item Xiangnan Li, \textit{Tongji University}

    \item Yuante Li, \textit{Tongji University}
    
    \item Yixing Li, \textit{The Chinese University of Hong Kong}

    \item Zhentao Lai, 
    \textit{Tongji University}
    
    \item Zilu Ma%, 
    %\textit{ }
    
    \item Yingrong Peng, \textit{Tsinghua University}
    
    \item Jiacheng Qian, \textit{Tongji University}

    \item Hao-Yu Sun, \textit{The University of Texas at Austin }
    
    \item Jianbo Sun, \textit{Kyushu University}
    
    \item Zirui Wang, \textit{Tongji University}
    
    \item Siwei Wu, \textit{Jilin University}

    \item Zian Wang, \textit{Tongji University}
    
    \item Bin Xu, \textit{University of Science and Technology of China}

    \item Jianghao Xu, \textit{Tongji University}
     
    \item Yiyang Yu, \textit{Tongji University}
    
    \item Zichuan Yang, \textit{Tongji University}

    \item Hongji Zha, \textit{Tongji University}
    
    \item Ruichong Zhang, \textit{Tsinghua University}

\end{itemize}

\end{document}